# Can GPT-4 learn to analyse moves in research article abstracts?

**Danni Yu**
Beijing Foreign Studies University

**Marina Bondi**
University of Modena and Reggio Emilia

**Ken Hyland**
University of East Anglia

## ABSTRACT

One of the most powerful and enduring ideas in written discourse analysis is that genres can be described in terms of the moves which structure a writer's purpose. Considerable research has sought to identify these distinct communicative acts, but analyses have been beset by problems of subjectivity, reliability and the time-consuming need for multiple coders to confirm analyses. In this paper we employ the affordances of GPT-4/Copilot to automate the annotation process by using natural language prompts. Focusing on abstracts from articles in four applied linguistics journals, we devise prompts which enable the model to identify moves effectively. The annotated outputs of these prompts were evaluated by two assessors with a third addressing disagreements. The results show that an 8-shot prompt was more effective than one using two, confirming that the inclusion of examples illustrating areas of variability can enhance GPT-4's ability to recognize multiple moves in a single sentence and reduce bias related to textual position. We suggest that GPT-4 offers considerable potential in automating this annotation process, when human actors with domain-specific linguistic expertise inform the prompting process.

## INTRODUCTION

Move analysis is an approach to text analysis used to investigate the organizational and rhetorical structure of written genres. Pioneered by Swales (1990), it is an essential component of his genre model, referring to the recognizable stages of particular institutional genres and the constraints on typical move sequences. For Swales (2004: 228), a move is 'a discoursal or rhetorical unit that performs a coherent communicative function in a written or spoken discourse', thus it helps frame a writer/speaker's rhetorical decision-making to meet the expectations of a particular community. Moves, then, are genre-dependent and perform both local purposes for the writer and collectively contribute to the communicative purpose of the genre itself. Move analysis has been one of the most productive approaches to texts in recent years with a plethora of papers attempting to describe the structural organization of academic genres as diverse as mathematics research articles (Kuteeva and McGarth 2015), 3-minute theses (Hu and Liu 2018), and student lab reports (Parkinson 2017).

But while analyzing schematic structures has proved an invaluable way of looking at texts, analysts have long been aware of the dangers of interpreting units as mono-functional (Hyland, 2002) and ignoring writers' complex purposes and 'private intentions' (Bhatia 1999). There is also the problem raised by Crookes (1986) of validating analyses to ensure they are not simply products of the analyst's intuitions. Because moves are a functional, rather than a formal unit, there are difficulties in identifying specific lexico-grammatical units as conveying particular rhetorical functions. Transitions from one move to another in a text are, of course, always motivated outside the text as writers respond to their social context and personal goals, but analysts have not always been convincingly able to identify the ways these shifts are explicitly signalled (for a discussion see for example Gray et al. 2020).

As a result, move analysis is a complex, subjective and time-intensive business and replication of such studies is hampered by the limited information typically provided on the process by researchers (Moreno and Swales 2018). As Casal and Kessler have recently observed:

> Methodological issues such as the number of annotators, the procedure for developing/modifying a framework, the unit of analysis (linguistic or rhetorical), inter-coder reliability, and the procedures for resolving disagreements are not uniformly reported. (Casal and Kessler 2024: 96).

Transparency regarding these issues can be potentially enhanced by LLMs such as GPT-4. To condition a model for a move annotation task, it is essential to create well-drafted prompts with explicit descriptions of the moves and clear criteria for the unit of analysis, which can then serve as protocols for replicating the study. Additionally, as the model can be consistently used to annotate the entire corpus, the results are less likely to be affected by low inter-coder reliability.

In this paper we seek to resolve some of these methodological issues by designing prompts which enable the freely available GPT-4 model on the Microsoft Copilot interface (formerly Bing chatbot) to automatically annotate research article abstracts.

## ARTICLE ABSTRACTS AND RHETORICAL MOVES

The need to easily access information to guide reading has led, in recent years, to the widespread use of structured abstracts in the life sciences, and particularly in clinical journals (Hartley 2014). Authors are encouraged to label their abstracts with pre-defined headings (such as Purpose, Design, Analysis, Findings), essentially mirroring the IMRD (Introduction-Method-Results-Discussion) structure of a full article. Structured abstracts have subsequently expanded to other areas such as engineering (e.g. Radix and Mohammed 2017) and the humanities and social sciences. Despite the relatively standardized structure of abstracts, researchers fail to concur on the number and names of abstract moves. Bhatia (1993) for example, identifies four moves while Swales and Feak (2012), Dos Santos (1996), and Hyland (2004) recognize five. The present study follows the latter three studies in recognizing the move schema shown in Table 1:

[TABLE 1 NEAR HERE]

## TOWARDS AUTOMATED MOVE ANNOTATION

Recent studies have shown the considerable potential of GPT models to recognize and annotate various phenomena such as stance (Gilardi et al., 2023), text categories (Kuzman et al., 2023), pragmatic units in speech acts (Yu et al. 2024) and speech functions in conversations (Ostyakova et al., 2023). In particular, Yu et al. (2024) found that GPT-4/Bing (now called Copilot) achieved accuracy approaching that of a human coder in annotating pragmatic units, including those without conventional formal markers, such as reasons for an apology. In this paper we aim to assess whether move annotation tasks can be conducted with GPT-4 by using prompts with examples representing moves.

## METHODS: PROMPTING FOR MOVE ANNOTATION IN ABSTRACTS

To assess whether rhetorical moves can be annotated automatically by an LLM, we went through a set of exploratory and confirmatory analyses (Roettger, 2019) using the GPT-4 model on Copilot in creative mode [1]. As Copilot is now freely available, researchers can use the same platform to replicate our analysis. Our approach had three main stages: 1) task setting; 2) prompt designing; and 3) output assessing. The stage of prompt designing comprises a set of exploratory tests, where we observed the impact of prompting factors on outputs, in order to design a prompt that could complete the annotation task effectively. Building upon these exploratory analyses, we concluded that the number of examples included in the prompt may affect the model's performance in the annotation task. To test this possibility and provide details of the annotation, we systematically assessed the outputs of two prompts with different numbers of input examples. Details are provided below.

### Task setting

The task we assigned to the LLM was to annotate rhetorical moves in RA abstracts written in English from applied linguistics. The corpus under investigation was composed of 180 abstracts collected from four leading journals in applied linguistics, which are *Language Learning*, *Applied Linguistics*, *TESOL Quarterly*, and *Language Teaching Research*. These abstracts are drawn from the most recent articles, as of December 2023. The corpus was divided into five sub-corpora according to the chronological order of the texts. Among these, S1, S2, S3, and S4, which contained 20 abstracts each (5 from each journal), were used in prompt design. These were used in the different phases of prompt design. S5, which contained 100 abstracts (20 from each journal), was used for assessing the LLM's annotation performance with different prompts.

The selection of Labels is crucial when setting an annotation task. We used simple, precise, and self-evident labels for our annotation task: BACKGROUND, PURPOSE, METHOD, RESULTS and CONCLUSION (Table 1). These labels are both explicit and succinct while not necessarily presupposing information about textual position associated with terms such as "introduction". We followed two criteria in this process: 1) the labels were assigned at sentence level; 2) a sentence with overlapping moves should be assigned multiple labels.

**Prompt design**

To prepare the LLM for the annotation task, we paid particular attention to prompt design. Prompting is the method of conditioning, or 'teaching', the model to generate expected outputs (Liu et al. 2023). In this study, prompt design involved two steps. The first aimed to draft a prompt that could help the model generate coherent and contextually relevant responses in appropriately completing the annotation task, at least on "easy cases", i.e., instances with explicit linguistic markers indicating the function of a move (e.g., *we conclude with recommendations for* which indicates the conclusion move). This step was conducted on S1. Here we consulted instructional articles on Microsoft Learn[2] related to prompting strategies. This meant, following, for example, suggestions such as *being specific, being descriptive by using analogies*, *repeating instructions*, *ordering matters to account for the model's recency bias*, and *giving the model an example of output to avoid generating false responses*[3]. Based on trials with S1, a candidate prompt was established, using the following strategies:

1) Using clear boundaries and signals to separate the different sets of instructions. The annotation task required multiple instructions, so we divided them into paragraphs and labelled them as instruction 1, instruction 2, etc. This helped the model to learn the instructions one by one.
2) Providing detailed definitions for each label. For instance, the METHOD move was explained with 'this move describes the research design, data collection, data analysis procedures, analysis techniques, and theories used in the study'. This descriptive definition helped the model identify the move more accurately than the simpler 'this move describes the methodology'.
3) Adapting the language to the model's own output. For example, we asked the model to define the five rhetorical moves and used the resulting expressions as a reference to refine the prompt.
4) Avoiding ambiguous indications that could confuse the LLM. For example, the LLM misinterpreted the indication 'annotate the following abstract with the five rhetorical moves' as meaning that an abstract must contain all five rhetorical moves. We resolved this by changing the instruction to 'annotate each sentence in the following abstract'.
5) Using explicit directives. For example, the model misunderstood the indication 'Learn instruction 1' as implying that we wanted to learn instruction 1. We corrected this misunderstanding by using the more explicit directive 'Please learn instruction 1'. This misunderstanding may occur because the LLM, unable to capture paralinguistic information, requires more explicit linguistic cues (*please* in this case) to comprehend the user's intent.
6) Emphasizing the aspects that the model often missed. For example, the model frequently overlooked combined moves. Therefore, we added the reminder 'pay attention to combined moves' in the instruction.

In the second step, the candidate prompt, which included directive indications, label definitions and a few annotated examples, was further refined through annotation trials with S1, S2, and S3. This step focused on developing instructions in the prompt to address specific annotation errors generated in the annotation trials. More specifically, we built an error corpus, which contained 'difficult cases' that the model failed to annotate correctly. Focusing on the error corpus, we refined the prompt by adding *ad hoc* instructions and relevant examples to enhance the model's performance in annotating these cases. Three error types emerged, and specific prompting strategies were undertaken to address them, as shown here:

[ERROR TYPES NEAR HERE]

**A pilot experiment**
To assess the improvement of the model's performance with different prompts, we conducted a pilot experiment on ten texts randomly selected from the error corpus, using nine prompts containing different numbers of examples (Table 2). These prompts, except for the 0-shot prompt, applied the *few-shot prompting*[4] technique, which exposes the model to examples ('shots') for in-context learning, a widely recognized ability of LLMs in learning from a few task-relevant demonstrations to improve their performance (Brown et al., 2020).

[TABLE 2 NEAR HERE]

We evaluated these annotations in terms of sentence-level accuracy, assessing the proportion of sentences correctly coded out of the total number of sentences. The results show that there was a stable improvement from the 0-shot prompt to the 4-shot prompt, but a drop emerged with the 5-shot and 6-shot prompts (Figure 1). The errors generated by these two prompts concern mainly the false recognition of the first sentence as BACKGROUND, which may be due to the fact that these two prompts have added examples beginning with BACKGROUND. To address this issue, in the 7-shot and 8-shot prompts we added examples that do not begin with BACKGROUND. The accuracy of the

8-shot prompt achieved 93.33%.

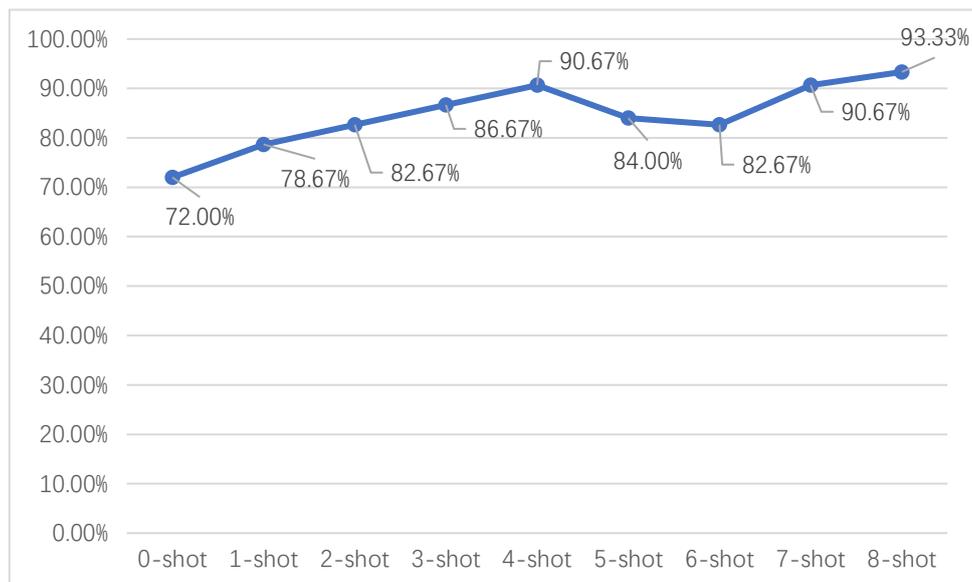

**Figure 1**. Accuracies of the annotations generated with the nine prompts on ten texts

The 8-shot prompt was composed of one instruction for label definition and eight instructions with examples illustrating various move patterns. In particular, in the last instruction, we asked the model to perform a trial annotation and gave it corrective feedback. As the Copilot chatbot had a limit of 4000 characters for each chat turn (at the time we conducted the experiments), we divided the prompt into four sets, each to be input separately.

Subsequently, we conducted another pilot experiment on the effectiveness of the 8-shot prompt on S4. Having noted that the model's performance might vary across different chat sessions, we carried out two rounds of annotation, employing the same prompt to annotate S4. This was done to assess if Copilot performed stably across two distinct tests. The assessment of a human evaluator showed that both tests reached sentence-level accuracy over 90% (Table 3). The high accuracy indicates the feasibility of applying the prompt in move annotation tasks. The efficacy of the prompt will be further tested in a larger sample, as will be presented in the next section.

[TABLE 3 NEAR HERE]

**Assessing output**
The comparison of the effectiveness of the 8-shot prompt and the 2-shot prompt was conducted on S5 composed of 100 texts with 678 sentences in total. Two human evaluators judged the accuracy of the LLM's annotated outputs. The results indicated a disagreement of 1.8% between the two human evaluators, with disagreement over just 12 GPT-annotated sentences. These different assessments concerned mainly the moves METHOD (six cases) and PURPOSE (six cases). A third assessor, with considerable expertise in academic discourse studies, was asked to review the 12 cases to decide whether the model correctly annotated a move. The final assessment results were measured using the parameters of *precision*, *recall*, and *F1-score*. To take the label METHOD as an example,

- *Precision* indicates the proportion of instances correctly assigned with METHOD among all the instances assigned with this move. It is calculated as the number of True Positives (TP) divided by the sum of True Positives and False Positives (FP). Precision = TP / (TP + FP).
- *Recall* indicates the proportion of instances correctly identified as METHOD among all actual instances of this move in the sample. It is calculated as the number of True Positives divided by the sum of True Positives and False Negatives (FN). Recall = TP / (TP + FN).
- *F1-score* is the harmonic mean between precision and recall and provides overall assessment of the model's accuracy in annotating METHOD. F1 Score = 2*(Precision*Recall) / (Precision + Recall).

To sum up, Table 4 provids an overview of the steps used in the stages of task setting, prompt design and assessment.

[TABLE 4 NEAR HERE]

**EXPERIMENTAL RESULTS AND ANALYSIS**
The outputs generated by both the 8-shot and 2-shot prompts have high accuracy readings (Table 5), indicating that GPT-4 can be used to assist rhetorical move annotation. The model accurately labelled all five moves with F1 over 0.8, except for PURPOSE which was annotated with the 2-shot prompt at 0.78 accuracy.

[TABLE 5 NEAR HERE]

Although the overall results for both prompts are impressive, we can observe a gap in the results in Table 5 with the 8-shot prompt performing better in identifying all the labels than the 2-shot prompt,. This suggests that the use of more examples and instructions can help condition the model to better understand the task. An example annotated by GPT-4 using the 8-shot prompt is shown in Figure 2.

> OK, I will try to annotate the abstract according to the instructions you have given me. Here is my attempt:
>
> <BACKGROUND>The lexicon of emotion words is fundamental to interpersonal communication.</BACKGROUND> <PURPOSE><METHOD>To examine how emotion word acquisition interacts with societal context, the present study investigated emotion word development in three groups of child Korean users aged 4–13 years: those who use Korean primarily outside the home as a majority language (MajKCs) or inside the home as a minority language (MinKCs), and those who use Korean both inside and outside the home (KCs).</METHOD></PURPOSE> <METHOD>These groups, along with a group of L1 Korean adults, rated the emotional valence of 61 Korean emotion words varying in frequency, valence, and age of acquisition.</METHOD> <RESULTS>Results showed KCs, MajKCs, and MinKCs all converging toward adult-like valence ratings by ages 11–13 years; unlike KCs and MajKCs, however, MinKCs did not show age-graded development and continued to diverge from adults in emotion word knowledge by these later ages.</RESULTS> <CONCLUSION>These findings support the view that societal context plays a major role in emotion word development, offering one reason for the intergenerational communication difficulties reported by immigrant families.</CONCLUSION>
>
> How did I do? Please give me some feedback on my annotation. If you want to end the conversation, please type "bye".

**Figure 2.** An example of output generated by GPT-4/Copilot with the 8-shot prompt.

In the following discussion we will focus on mislabelled examples and explore possible solutions to further improve the accuracy of the move-analysis task.

**Over-recognition of BACKGROUND: meanings concealed by textual position**

We can see from Table 5 that among all the labels, BACKGROUND led to the lowest F1 score for both prompts. The precision score is particularly low (0.81 with the 8-shot prompt and only 0.69 with the 2-shot prompt), while the recall score is high with both prompts (0.93 vs 0.94). This means that the model was able to recognize most of the true BACKGROUND moves, but had a tendency to excessively identify other moves as BACKGROUND. With a qualitative analysis of the cases of false BACKGROUND, we noticed that the misrecognition of this move seemed to be related to sentence location in the text. The model tended to over-recognize the first sentence as BACKGROUND while ignoring the more prominent presence of other moves such as METHOD and PURPOSE. This phenomenon where certain meanings are concealed by the textual position occurred 14 times in the experiment with the 8-shot prompt and 29 times with the 2-shot prompt. An example of over-recognized BACKGROUND can be found in Figure 3.

[FIGURE 3 NEAR HERE]

In our assessment, the sentence labelled as BACKGROUND was considered a combination of METHOD (*draws on the concept*) and PURPOSE (*to interpret the silence*). While BACKGROUND and METHOD may somehow overlap, we think that the sentences match our definitions of METHOD (including theory) combined with PURPOSE. However, the model seemed to prioritize what appeared to be the position-related meaning of BACKGROUND. Lexico-grammatical analysis of the cases of false BACKGROUND revealed a frequent presence of markers of PURPOSE (and METHOD), such as *this article examines*, *this study focuses on*, *this mixed-methods study investigates*, *this article reports on*, and *this study explores*. These explicit PURPOSE markers seemed to be "overshadowed" by BACKGROUND when they appeared in the first sentence of an abstract.

**Recognition of PURPOSE and METHOD: challenges with combined moves**
In abstracts, the PURPOSE and METHOD moves are often merged (e.g. Dos Santos 1996). This tendency was also noted in our corpus, posing a significant challenge for the model. The results show that the 2-shot prompt was much less effective in recognizing these two moves in combination compared to the 8-shot prompt (Table 5). Upon examining the data, we identified two main cases where PURPOSE and METHOD were ignored by the model. The first case concerns the misrecognition of BACKGROUND in a sentence which should have been annotated with PURPOSE and/or METHOD (Figure 3). The second case concerns the model's failure to recognize combined moves. Using the 2-shot prompt, the model tended to assign only one label to a sentence that should have been annotated with two moves, omitting PURPOSE or METHOD. For example:

(1) <METHOD>**It also introduces a novel method** of examining developmental trajectories that uses both inferential statistics and descriptive measures **to account not only for relationships between the year of study and use of linguistic features, but also for the shape of the trajectories and frequencies of occurrence over time.**</METHOD> (Unrecognized PURPOSE, with the 2-shot prompt)
(2) <PURPOSE>This study compared auditory and orthographic presentations of novel words with different degrees of phonological overlap **using CSWL in a laboratory-based and an online-based approach**.</PURPOSE> (Unrecognized METHOD, with the 2-shot prompt)

The relative inability of the 2-shot prompt to recognize combined moves may be due to the fact that only one example with combined moves was provided in the prompt. The 8-shot prompt, on the other hand, was provided with eight examples annotated with combined moves, which we consider a useful strategy in enhancing the model's ability to recognize this phenomenon.

**Recognition of RESULTS: being overshadowed by a leading move**
The RESULTS move achieved high F1 score with both prompts. However, in terms of recall, the 8-shot prompt appeared to be more effective than the 2-shot prompt. The former failed to identify 6 cases out of the 219 instances of RESULTS, while the latter missed 15 cases. Analyzing the unrecognized RESULTS, we noticed that the move, although expressed with explicit markers such as *revealed that*, *illustrates*, *the findings*

*evince*, *our findings show*, tended to be neglected when it was 'overshadowed' by the leading move, mostly METHOD, that initiated the sentence, as here:

(3) <METHOD>Analysis of data from classroom observations and semi-structured interviews with the teacher **revealed** three major LOA implementation endeavors in response to the challenges she faced:</METHOD> (Unrecognized RESULTS, with both the 2-shot and 8-shot prompts)
(4) <METHOD>Through Frame Analysis, **the findings evince** how these language learners challenge their usual linguistic and masculine habitus and how they achieve interesubjectivities through the symbolic powers of English, Cypriot Greek, and Standard Greek.</METHOD> (Unrecognized RESULTS, with both the 2-shot and 8-shot prompts)

This error represents also one reason for the unrecognition of combined moves. The greater effectiveness of the 8-shot prompt might be accounted for by the inclusion of an example annotated with METHOD and RESULTS. To further enhance the model's performance for this case, we could add an example like example 5 or 6 to the prompt, which would indicate that sentences starting with METHOD can also have other moves.

**Recognition of CONCLUSION: a case of hallucination**
The model performed outstandingly in recognizing CONCLUSION. Among the 94 cases of CONCLUSION in the corpus, only one was ignored by the 8-shot prompt, while two cases were missed by the 2-shot prompt. The precision was also exceptionally high. No case was falsely predicted with the 8-shot prompt, while the 2-shot prompt falsely identified only 3 cases.

**Summing up**
The errors discussed above can be categorized into three major types: unrecognition of a move, false recognition of a move, and hallucination (Table 6). Additional analysis of instances representing these errors reveal possible relevant factors related to them.

[TABLE 6 NEAR HERE]

Building on the comparison between the 2-shot and 8-shot prompts in terms of error types, we have discussed the pros and cons of the two prompts and offered potential countermeasures to address some of the errors. In particular, the enhanced performance of the 8-shot prompt in recognizing combined moves and in reducing the over-recognition of BACKGROUND indicates that the inclusion of examples illustrating relevant linguistic phenomena could help the model more effectively manage these challenges. However, the emphasis on a particular linguistic phenomenon could possibly induce the model to over-recognize that phenomenon. For example, the use of the 8-shot prompt has led to a higher frequency of over-annotation with combined moves compared to the 2-shot prompt. This suggests that we need to determine priorities when using this automated analysis .

**CONCLUSION**

The present study provides evidence to the applicability of Large Language Models such as GPT-4 in assisting rhetorical move annotation through skillful prompting. The results show that the 8-shot prompt was more effective than the 2-shot prompt, which suggests that the inclusion of examples, illustrating variations in rhetorical move structures, can contribute to enhancing the overall accuracy of the assigned task.

Perhaps equally importantly, we should point out that one result of our study is the conclusion that, in the era of Artificial Intelligence and the use of LLM-driven bots for language analysis, domain-specific linguistic expertise is essential for appropriate prompt design. As illustrated in this study, for the task of rhetorical move annotation, expertise in linguistics, or more specifically, in corpus-based genre analysis, can aid in at least the following six steps: 1) collection of representative texts of the genre under investigation; 2) identification of moves in the genre; 3) establishment of a move scheme containing adequate definitions of each move; 4) selection of representative examples annotated with moves to be used in the prompt (representative examples may emerge during the tests); 5) assessment of the annotated results, paying particular attention to inconsistent cases (which may offer opportunities for theory-building); 6) qualitative analysis of incorrectly annotated cases to find possible solutions to enhance the accuracy of the model.

The GPT-assisted move annotation approach offers significant practical advantages for genre studies. First, and most importantly, this automated approach, with human verification, can produce faster and more objective results than those relying entirely on human annotation. Second, the approach can be effectively integrated into writing classrooms. For example, teachers can encourage students to devise prompts that guide a GPT interface in annotating rhetorical moves within texts of a specific genre. This process engages students in the six steps mentioned above and can develop their understanding of generic move functions, thus allowing them to improve their own writing. While there remains some way to go in exploiting the affordances of AI in language analysis, we believe this is a valuable beginning and involves genre analysts in a fascinating area of development.

Yu, D., L. Li, H. Su and M. Fuoli. 2024. 'Assessing the potential of LLM-assisted annotation for corpus-based pragmatics and discourse analysis: The case of apology,' *International Journal of Corpus Linguistics*.

APPENDIX

## Appendix 1. The 8-shot prompt

(refer to the forthcoming published version)

---

[1] There are three different chat tones available for Copilot in Windows: the creative mode, which gives responses that are longer and more descriptive; the precise mode, which focuses on shorter, more search-focused answers; and the balanced mode, which is somewhere in-between (see https://support.microsoft.com/en-us/windows/welcome-to-copilot-in-windows-675708af-8c16-4675-afeb-85a5a476ccb0).

[2] https://learn.microsoft.com/

[3] https://learn.microsoft.com/en-us/azure/ai-services/openai/concepts/prompt-engineering

[4] A few-shot prompt is a type of input that provides several examples to the model, enabling it to handle complex tasks. Other types of inputs are zero-shot prompts, giving no examples, and one-shot prompts, with only one example.